\documentclass{article}

\usepackage{PRIMEarxiv}

\usepackage[utf8]{inputenc} 
\usepackage[T1]{fontenc}    
\usepackage{hyperref}       
\usepackage{url}            
\usepackage{booktabs}       
\usepackage{amsfonts}       
\usepackage{nicefrac}       
\usepackage{algorithmic}
\usepackage[caption=false,font=normalsize,labelfont=sf,textfont=sf]{subfig}
\usepackage{cite}
\usepackage{algorithm}
\usepackage{array}
\usepackage{textcomp}
\usepackage{stfloats}
\usepackage{amsmath}
\usepackage{color}
\usepackage{xcolor}
\usepackage{orcidlink}
\usepackage{multirow}
\usepackage{microtype}      
\usepackage{lipsum}
\usepackage{fancyhdr}       
\usepackage{graphicx}       
\graphicspath{{media/}}     

\title{TextDestroyer: A Training- and Annotation-Free Diffusion Method for Destroying Anomal Text \\
from Images
}

\author{
  Mengcheng Li \\
  Key Laboratory of\\
  Multimedia Trusted Perception\\
  and Efficient Computing\\
  Xiamen University \\
  Xiamen, China\\
  \texttt{limengcheng@stu.xmu.edu.cn} \\
   \And
  Fei Chao \\
  Key Laboratory of\\
  Multimedia Trusted Perception\\
  and Efficient Computing\\
  Xiamen University \\
  Xiamen, China\\
  \texttt{fchao@xmu.edu.cn} \\
}

\begin{document}
\maketitle

\begin{abstract}
In this paper, we propose TextDestroyer, the first training- and annotation-free method for scene text destruction using a pre-trained diffusion model. Existing scene text removal models require complex annotation and retraining, and may leave faint yet recognizable text information, compromising privacy protection and content concealment. 
TextDestroyer addresses these issues by employing a three-stage hierarchical process to obtain accurate text masks. Our method scrambles text areas in the latent start code using a Gaussian distribution before reconstruction. During the diffusion denoising process, self-attention key and value are referenced from the original latent to restore the compromised background. Latent codes saved at each inversion step are used for replacement during reconstruction, ensuring perfect background restoration.
The advantages of TextDestroyer include: (1) it eliminates labor-intensive data annotation and resource-intensive training; (2) it achieves more thorough text destruction, preventing recognizable traces; and (3) it demonstrates better generalization capabilities, performing well on both real-world scenes and generated images.
\end{abstract}


\section{Introduction}
\begin{figure*}[!t]
    \centering
    \includegraphics[width=\textwidth]{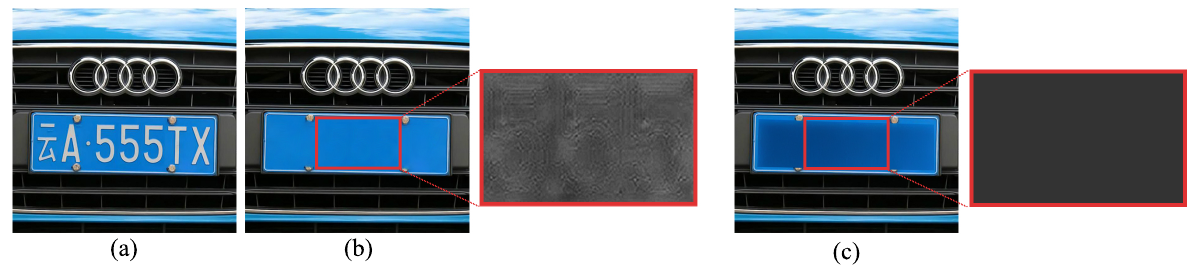}
    \caption{(a) An example of privacy text in an image: a photo of the front of a car, featuring the vehicle's license plate information. (b) The text removal method, DeepEraser~\cite{feng2024deeperaser}, still exposes the digital information ``555,'' despite its improved background recovery capabilities. (c) Using our TextDestroyer, the text information is entirely obliterated. Best view with zooming in.
    }
    \label{fig:motivation}
\end{figure*}

Recently, diffusion models~\cite{dhariwal2021diffusion,ho2020denoising,song2020denoising,rombach2022high,saharia2022photorealistic,balaji2022ediff,deepfloyd-if,esser2024scaling} have made remarkable achievements in text-conditioned image generation, enabling users to effortlessly transform their vivid imaginations into reality. 
They have shown impressive success in accurately and coherently rendering textual content. 
%
In particular, with the use of T5 text encoder component~\cite{raffel2020exploring}, preliminary text generation capabilities have been demonstrated by Imagen~\cite{saharia2022photorealistic}, eDiff-I~\cite{balaji2022ediff}, and DeepFloyd-IF~\cite{deepfloyd-if}. Liu \textit{et al.}~\cite{liu2022character} further improved text generation by employing character-aware text encoders~\cite{xue2022byt5}. To provide more precise guidance in text generation through diffusion, a series of efforts~\cite{chen2024textdiffuser,ma2023glyphdraw,yang2024glyphcontrol,zhang2024brush} have been dedicated to designing specialized network architectures for generating refined text. Stable Diffusion 3~\cite{esser2024scaling} abandons the traditional U-Net architecture in favor of DiT~\cite{peebles2023scalable} for denoising. In addition to producing awe-inspiring images, it also shows the ability to accurately represent textual information within the images.
Easy access to tools that create images from text can lead to problems with copyright, privacy, and the law. Sharing personal information like phone numbers or addresses in images online has already been a big issue, as shown in Fig.~\ref{fig:motivation}(a). People try to hide this information by making it blurry or covering it up, but this leads to unintended complications.
With more powerful image-making tools, these issues could get worse. Bad people might use these tools to spread false information by putting sensitive details in fake scenes. It's important to think about both the good and bad sides, especially how to stop the spread of unwanted text in images.

In line with our motivation, the field of scene text removal has been extensively studied. Before the era of deep learning, numerous efforts had already employed traditional machine learning and computer vision techniques to tackle this task~\cite{pnevmatikakis2008inpainting,khodadadi2012text,modha2014image,wagh2015text}. 
Neural networks, with their powerful learning capabilities, have further enhanced the effectiveness of text erasure. 
Scene Text Eraser~\cite{nakamura2017scene} trained an end-to-end CNN, selectively erasing text from images divided into multiple patches. 
EnsNet~\cite{zhang2019ensnet} features a more sophisticated network structure for erasing both text and other objects.
MTRNet~\cite{tursun2019mtrnet}, using a conditional generative adversarial network (cGAN) with an auxiliary mask, further improved the effectiveness of erasure. 
PERT~\cite{wang2021pert} embeds a detection branch within the network, providing explicit guidance for erasure. 
DeepEraser~\cite{feng2024deeperaser} adopts a recursive architecture to gradually remove text over multiple iterations. 
Nonetheless, as depicted in Fig.~\ref{fig:motivation}(b), removal models may excessively focus on background restoration accuracy, potentially leaving subtle residues that enable the text to remain readable in Fig.~\ref{fig:motivation}(b).
Furthermore, these methods necessitate training, consuming considerable computational resources. Additionally, the training datasets demand labor-intensive manual annotation of masks and ground-truth images post-inpainting. 
Although STRDD~\cite{yang2022strdd} recently used diffusion for text removal, it still requires retraining and data annotation.
We recognize that by using a pre-trained diffusion model, the necessity for retraining or data annotation in text destruction can be circumvented. This is due to two factors: first, during the pre-training phase, diffusion models have already been exposed to numerous images, including those with scene text. Second, the diffusion architecture's cross-attention mechanism exhibits rough localization capabilities for textual regions in latent space, providing an alternative to manual mask annotation. Thus, it becomes feasible to explore a training-free and annotation-free text deconstruction method.

In this paper, we introduce TextDestroyer, the first training- and annotation-free diffusion method for scene text destruction. We emphasize training-free techniques, automatic text localization, and comprehensive destruction of textual regions. 
%
%
We then employ a hierarchical process for progressive and precise text localization. 
In the \textit{introductory text capturing} stage, we aggregate multiple token-level attention maps from the inversion process and segment them to capture an introductory text region mask. 
In the \textit{continuous text adjustment} stage, we crop and resize all text regions in the original image and apply the same inversion process to adjust text regions with reduced background interference.
In the \textit{meticulous text delineation} stage, we perform 2-means clustering on the original image, using the non-text areas from the second stage as a reference to distinguish between text and background clusters. 
With a precise mask of text areas, we destroy their latent codes using random Gaussian noise before reconstructing the image through the diffusion denoising process.
Also, we introduce a denoising process to guide image reconstruction, replacing the erroneous latent codes with original ones at each step for low distortion of background. 
This diffusion process offers key $K$ and value $V$ of non-text areas at specific time steps and self-attention layers for denoising reconstruction, enabling background restoration. To further ensure low distortion in non-text areas, we replace the latent code during the reconstruction when denoising is nearly complete.
Finally, we accomplish a complete obliteration of scene text as illustrated in Fig.~\ref{fig:motivation}(c).

The major contributions of this paper are summarized as:

1. {\bf Training and Annotation-Free Approach}. TextDestroyer is the first method to destroy scene text without requiring additional training or annotations. This approach simplifies the text removal process, as it does not rely on labor-intensive data annotation or resource-intensive model training, making it efficient and accessible for practical applications.

2. {\bf Enhanced Text Destruction and Background Restoration}. The method employs a three-stage hierarchical process that not only ensures thorough destruction of text but also enhances background restoration. It uses Gaussian noise to scramble text regions and a diffusion denoising process for image reconstruction, which preserves the background integrity by replacing erroneous latent codes with original ones during reconstruction, minimizing visual distortions and maintaining the quality of non-text regions.
\section{Related Work}

\subsection{Scene Text Removal}
Scene text removal involves erasing text information from real images and filling the erased areas with content similar to the remaining portions. Early research employed non-learning-based methods~\cite{pnevmatikakis2008inpainting,khodadadi2012text,modha2014image,wagh2015text} for text erasing, using color-histogram- or threshold-based techniques to locate text regions and similarity-based smoothing methods for inpainting.
Recent studies applied deep learning-based methods~\cite{liu2022don,yang2022strdd,bian2022scene,feng2024deeperaser,nakamura2017scene,zhang2019ensnet,liu2020erasenet,wang2021pert} to scene text removal. Several studies~\cite{tursun2019mtrnet,lee2022surprisingly,liu2022don,yang2022strdd,bian2022scene,feng2024deeperaser} maintain separate stages for localization and erasure, while others integrate the erasure into an end-to-end model~\cite{nakamura2017scene,zhang2019ensnet,liu2020erasenet,wang2021pert}, reducing the difficulty of data collection and training. However, these pipelines and end-to-end models still require training, and their highly specialized structures make it challenging to transfer or extend to other tasks.
Diffusion models have been applied~\cite{yang2022strdd}, serving as a black box, limiting its potential to locate image content.

\subsection{Diffusion Editing}
Diffusion models~\cite{dhariwal2021diffusion,ho2020denoising,song2020denoising,rombach2022high,esser2024scaling} initiate from Gaussian noise and generate images through random denoising steps. DDPM~\cite{ho2020denoising} uncovers images from noise during Markovian reverse processes. DDIM~\cite{song2020denoising} denoising on non-Markovian processes reduces sampling steps. LDMs~\cite{rombach2022high} transition to the latent space, operating at a reduced resolution. Several studies~\cite{meng2021sdedit,nichol2022glide,avrahami2022blended,cao2023masactrl,ruiz2023dreambooth,kawar2023imagic} have leveraged the generative capacity of diffusion to develop real image editing, targeting the removal and replacement of undesirable image components. Some works refine the diffusion model~\cite{ruiz2023dreambooth,kawar2023imagic} or its trainable counterpart~\cite{zhang2023adding} to enhance control over the edited object's structure and texture. However, fine-tuning diffusion models is resource-intensive. Many studies~\cite{cao2023masactrl} utilize feature map to control shape and texture without tuning. Models based on LDMs employ pre-trained language models like CLIP~\cite{radford2021learning} as text encoders, infusing conditions into the U-Net, thus aligning prompts with image semantics. 
Liu \textit{et al.}~\cite{liu2022character} pointed out that character-blind and capacity-limited diffusion models face challenges in perceiving text regions. Many priors on text perception and editing, such as fine-tuning~\cite{yang2024glyphcontrol}, external components~\cite{chen2024textdiffuser}, or user-provided masks~\cite{chen2024textdiffuser}, increase computational demands and impact user experience.


\section{Methodology}

\begin{figure*}[!t]
    \centering
    \includegraphics[width=\textwidth]{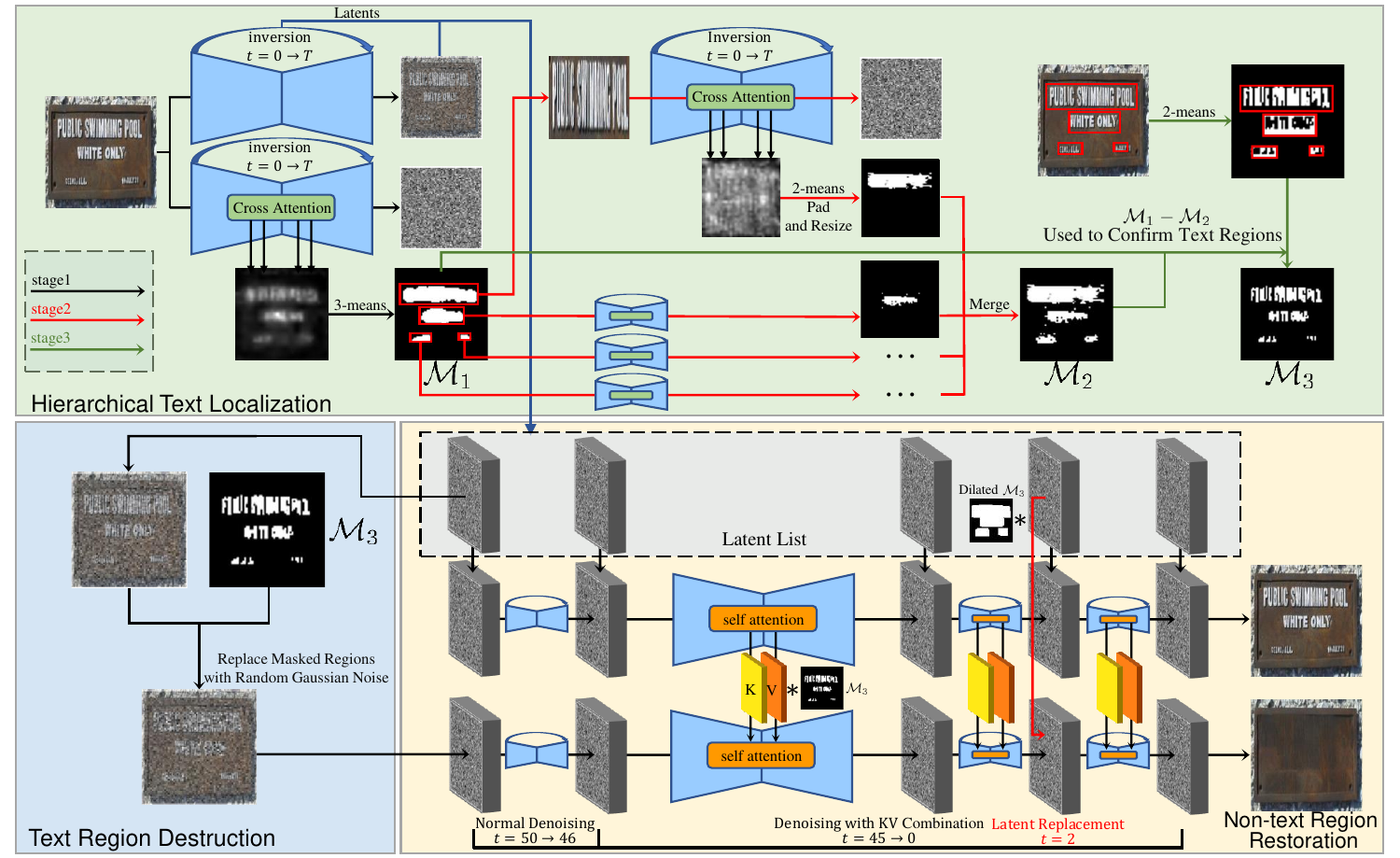}
    \caption{Overall framework of proposed TextDestroyer. The top manifests its three-stage hierarchical text localization to capture an introductory text area $\mathcal{M}_1$, a continuously adjusted text area $\mathcal{M}_2$, and finally the meticulous text boundaries $\mathcal{M}_3$. Bottom left displays its text region destruction by replacing text boundaries with random Gaussian noise. Bottom right shows its non-text region restoration in a fashion of $KV$ combination and latent replacement.}
    \label{fig:framework}
\end{figure*}

In this section, we formally introduced our TextDestroyer method, framework of which is provided in Fig.~\ref{fig:framework}.

\subsection{Preliminaries}
\subsubsection{Latent Diffusion Models}
%
%

Latent diffusion models (LDMs) employ an autoencoder $\mathcal{E}$ that encodes an image $\boldsymbol{x}_{0} \in \mathbb{R} ^ {H \times W \times 3}$ into a low-dimensional latent space $\boldsymbol{z}_{0} = \mathcal{E}(\boldsymbol{x}_{0}) \in \mathbb{R} ^ {h \times w \times c}$. Here, $f = H/h = W/w$ represents the downsampling factor, and $c$ denotes the channel dimension. The forward diffusion process is defined as:
\begin{equation}
\boldsymbol{z}_t = \sqrt{\bar{\alpha}_t} \boldsymbol{z}_0 + \sqrt{1-\bar{\alpha}_t}\epsilon, \quad \epsilon \sim \mathcal{N}(\boldsymbol{0}, \boldsymbol{I}),
\end{equation}
where ${\{\alpha_t\}}_{t=1}^{T}$ represents a set of predetermined variance schedules, and $\bar{\alpha}_t = \Pi_{i=1}^t \alpha_i$. A U-Net $\epsilon_{\theta}$ serves as a conditional denoiser, which estimates noise incrementally to recover the image's latent representation $\boldsymbol{z}_0$ from the random Gaussian noise $\boldsymbol{z}_T$:
\begin{equation}\label{eq:denoise}
\boldsymbol{z}_{t-1} = 
\sqrt{\frac{\alpha_{t-1}}{\alpha_{t}}} \boldsymbol{z}_{t} + 
(\sqrt{\frac{1}{\alpha_{t-1}}-1}-\sqrt{\frac{1}{\alpha_{t}}-1})\epsilon_{\theta}
(\boldsymbol{z}_{t},t,\tau_{\theta}(\mathcal{P})),
\end{equation}
where $\tau_{\theta}(\mathcal{P})$ denotes a text encoder that converts the conditional text prompt $\mathcal{P}$ into an embedding. The conditional embedding $\tau_{\theta}(\mathcal{P})$ and the intermediate representation of noise in the U-Net $\phi(\boldsymbol{z}_{t})$ are combined through attention computation in cross-attention (CA) layers. This integration introduces information from the user-specified text prompt into the U-Net's generation process:
\begin{equation}\label{eq:crossattn}
\begin{split}
Q=W_{Q}\cdot\phi(\boldsymbol{z}_{t}),\;  K=&W_{K}\cdot\tau(\mathcal{P}),\;  V=W_{V}\cdot\tau(\mathcal{P}),\\
CA(Q,K,V)=&Softmax(\frac{QK^T}{\sqrt{d}})\cdot V.
\end{split}
\end{equation}

At time step $t=0$, a decoder $\mathcal{D}$ decodes the latent space output $\boldsymbol{z}_{0}$ into the high-dimensional pixel space $\boldsymbol{x}_{0} = \mathcal{D}(\boldsymbol{z}_{0})$.

\subsubsection{DDIM Inversion}

Utilizing DDIM sampling~\cite{song2020denoising}, a deterministic sampling process can be attained by fixing the variance per Eq.~\eqref{eq:denoise}.
Under the assumption that the ordinary differential equation (ODE) process is reversible with small steps, the DDIM sampling enables an inversion process to facilitate the transition from $\boldsymbol{z}_{0}$ to $\boldsymbol{z}_{T}$, which can be formulated by the following equation: 
%
%
\begin{align}\label{eq:inversion}
\boldsymbol{z}_{t}^{*} = 
&\sqrt{\frac{\alpha_{t}}{\alpha_{t-1}}}\boldsymbol{z}_{t-1}^{*}\: + \notag \\
&\sqrt{\alpha_{t}} \left( \sqrt{\frac{1}{\alpha_{t}}-1} - \sqrt{\frac{1}{\alpha_{t-1}}-1} \right)\epsilon_{\theta}(\boldsymbol{z}_{t-1}^{*}, t-1, \tau_{\theta}(\mathcal{P})).
\end{align}

By initiating the process with $\boldsymbol{z}_{T}^{*}$ and continuing the denoising by Eq.~\eqref{eq:denoise}, we can obtain an approximate $\boldsymbol{z}_{0}^{*}$ of the original latent $\boldsymbol{z}_{0}$. 
Our major objective is thus to create $\boldsymbol{z}_{0}^{*}$ without text information compared to the original $\boldsymbol{z}_{0}$.
%
%


\subsection{Hierarchical Text Localization}\label{sec:localization}

\begin{algorithm}[!t]
    \caption{Hierarchical Text Localization}
    \label{alg:localization_standard}
    \begin{algorithmic}[1]
        \REQUIRE input image $\boldsymbol{x}_{0}$.
        \ENSURE latent list $Z$ and accurate text mask $\mathcal{M}_{3}$.
        \STATE $\boldsymbol{z}_{0}=\text{autoencoder}(\boldsymbol{x}_{0})$
        \STATE \textbf{// Stage-1 and latent acquisition}
        \STATE Initialize arrays $M_1$ of size $T$ and $Z$ of size $T+1$
        \FOR{$t = 0$ to $T-1$}
            \STATE $M_{tmp}[0 \dots n_{tokens}-1],\_= \text{inversion}_{\theta} (\boldsymbol{z}_t,t,\tau_{\theta}(\mathcal{P}))$
            \STATE $\_,\boldsymbol{\epsilon}= \text{inversion}_{\theta} (\boldsymbol{z}_t,t)$
            \STATE $\boldsymbol{z}_{t+1} = 
\sqrt\frac{\alpha_{t+1}}{\alpha_{t}}\boldsymbol{z}_{t} + 
\sqrt{\alpha_{t+1}}(\sqrt{\frac{1}{\alpha_{t+1}}-1}-\sqrt{\frac{1}{\alpha_{t}}-1})\boldsymbol{\epsilon}$
            \STATE $M_1[t],Z[t] \leftarrow M_{tmp},\boldsymbol{z}_{t}$
        \ENDFOR
        \STATE $Z[T] \leftarrow \boldsymbol{z}_{T}$
        \STATE $M_{1}^*= \text{mean\_and\_aggregation} (M_{1})$
        \STATE $\mathcal{M}_{1}= \text{2-means\_segmentation} (M_{1}^*)$

        \STATE \textbf{// Stage-2}
        \STATE $boxes[0 \dots n_{boxes}-1] = \text{connected\_components}(\mathcal{M}_{1})$
        \STATE $X_{cropped}[0 \dots n_{boxes}-1] = \text{image\_cropping}(\boldsymbol{x}_{0},boxes)$
        \STATE Initialize an array $M_2$ of size $\text[n_{boxes},T]$
        \FOR{$i = 0$ to $n_{boxes}-1$}
            \STATE $\boldsymbol{z}_{0}=\text{autoencoder}(X_{cropped}[i])$
            \FOR{$t = 0$ to $T-1$}
                \STATE $M_{tmp}[0 \dots n_{tokens}-1],\boldsymbol{\epsilon}= \text{inversion}_{\theta} (\boldsymbol{z}_t,t,\tau_{\theta}(\mathcal{P}))$
                \STATE $\boldsymbol{z}_{t+1} = 
\sqrt\frac{\alpha_{t+1}}{\alpha_{t}}\boldsymbol{z}_{t} + 
\sqrt{\alpha_{t+1}}(\sqrt{\frac{1}{\alpha_{t+1}}-1}-\sqrt{\frac{1}{\alpha_{t}}-1})\boldsymbol{\epsilon}$
                \STATE $M_2[i,t] \leftarrow M_{tmp}$
            \ENDFOR
            \STATE $M_{2}^*[i]= \text{mean\_and\_aggregation} (M_{2}[i])$
            \STATE $\mathcal{M}_{2}[i]= \text{3-means\_segmentation} (M_{2}^*[i])$
        \ENDFOR
        \STATE $\mathcal{M}_{2}= \text{union} (\mathcal{M}_{2}[0],\mathcal{M}_{2}[1] \dots )$
        \STATE $\mathcal{M}_{2}= \text{dilation} (\mathcal{M}_{2},(k_1,k_1))$

        \STATE \textbf{// Stage-3}
        \STATE Initialize an array $\mathcal{M}_{3}$ of size $n_{boxes}$
        \STATE $\mathcal{M}_{ref}=\mathcal{M}_{1}-\mathcal{M}_{2}\ \; \text{//} \; \text{identify the background}$
        \FOR{$i = 0$ to $n_{boxes}-1$}
            \STATE $\mathcal{M}_{3}[i]= \text{2-means\_segmentation} (X_{cropped}[i],\mathcal{M}_{ref})$
        \ENDFOR
        \STATE $\mathcal{M}_{3}= \text{union} (\mathcal{M}_{3}[0],\mathcal{M}_{3}[1] \dots )$
        \STATE $\mathcal{M}_{3}= \mathcal{M}_{3} \cdot \mathcal{M}_{1}$
    \end{algorithmic}
\end{algorithm}

We have developed a hierarchical text localization process to accurately identify text regions for destruction, refining outlines incrementally until precise text edges are determined. This process, shown in Fig.~\ref{fig:framework}, consists of three stages. 
In the first stage, the average cross-attention map generated during the inversion process is sliced to capture an introductory text area $\mathcal{M}_1$.
In the second stage, each captured text area is cropped and magnified from the original image, to continuously adjust better text areas $\mathcal{M}_2$.
In the final stage, the original image undergoes a two-means clustering analysis, finally delineating the meticulous text boundaries $\mathcal{M}_3$.

\subsubsection{Introductory Text Capturing}
%
\begin{algorithm}[!t]
    \caption{Hierarchical Text Localization (Mask)}
    \label{alg:localization_mask}
    \begin{algorithmic}[1]
        \REQUIRE input image $\boldsymbol{x}_{0}$ and coarse text mask $\mathcal{M}_{user}$.
        \ENSURE latent list $Z$ and accurate text mask $\mathcal{M}_{3}$.
        \STATE $\boldsymbol{z}_{0}=\text{autoencoder}(\boldsymbol{x}_{0})$
        \STATE \textbf{// Latent acquisition}
        \STATE Initialize array $Z$ of size $T+1$
        \FOR{$t = 0$ to $T-1$}
            \STATE $\boldsymbol{\epsilon}= \text{inversion}_{\theta} (\boldsymbol{z}_t,t)$
            \STATE $\boldsymbol{z}_{t+1} = 
\sqrt\frac{\alpha_{t+1}}{\alpha_{t}}\boldsymbol{z}_{t} + 
\sqrt{\alpha_{t+1}}(\sqrt{\frac{1}{\alpha_{t+1}}-1}-\sqrt{\frac{1}{\alpha_{t}}-1})\boldsymbol{\epsilon}$
            \STATE $Z[t] \leftarrow \boldsymbol{z}_{t}$
        \ENDFOR
        \STATE $Z[T] \leftarrow \boldsymbol{z}_{T}$
        \STATE $\mathcal{M}_{1} \leftarrow \mathcal{M}_{user}$

        \STATE \textbf{// Stage-2 and stage-3 are consistent with Algorithm~\ref{alg:localization_standard}}
        \STATE $\dots$
    \end{algorithmic}
\end{algorithm}
\begin{figure*}[!t]
    \centering
    \includegraphics[width=\textwidth]{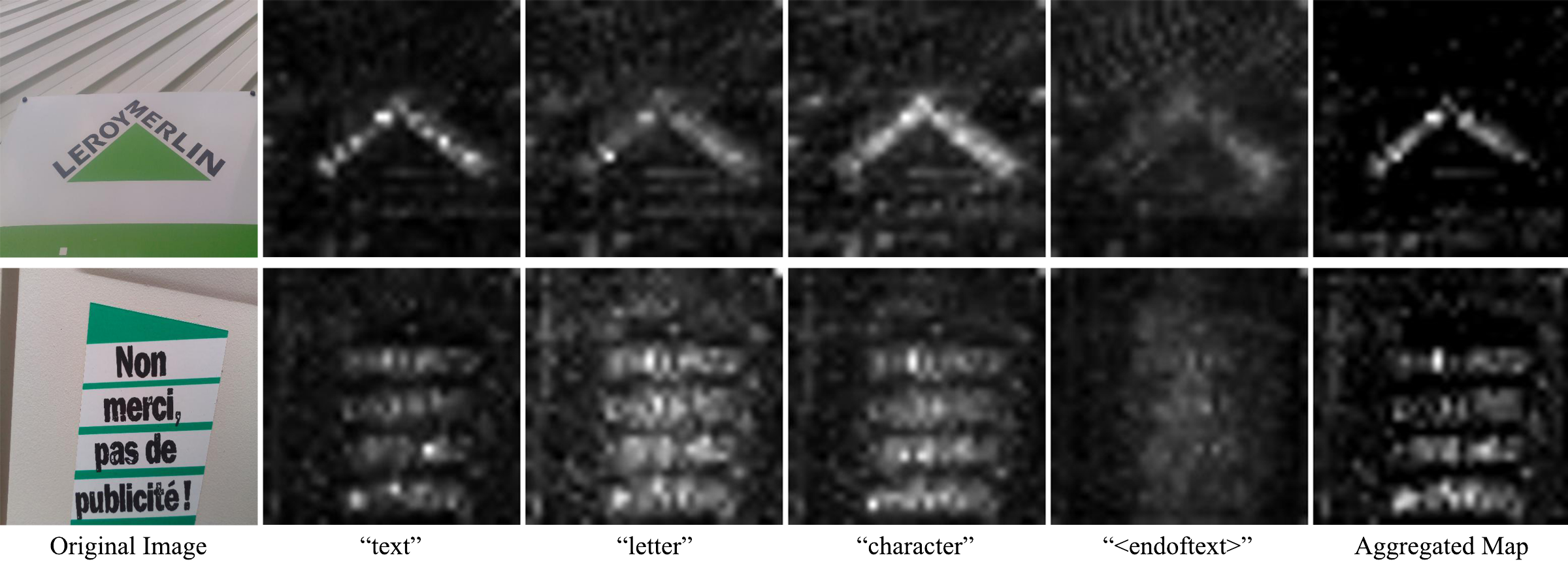}
    \caption{Token-level attention maps visualization.}
    \label{fig:attntionmaps}
\end{figure*}

Due to the limited latent resolution of stable diffusion, \emph{e.g.}, $64 \times 64$ for 1.5, encoding the entire image may cause the text region to occupy only a few pixels. 
Therefore, our introductory objective is to roughly capture the text area.

We make $\mathcal{P}$ = ``text letter character'' as the conditional prompt during inversion of Eq.~\eqref{eq:inversion} to capture the text areas.
The ``text'', ``letter'' and ``character'' ensure comprehensive coverage of text areas.
We compute $M^{*}=Q^{*}K^{{*}^{T}}$ from cross-attention layers during the \textit{inversion process}, following existing studies~\cite{hertz2022prompt,zhang2024brush} to yield a set of token-level attention maps $\{ M^{*}_{token} \}$ where $token$ = \{``text'', ``letter'', ``character''\}.
Fig.~\ref{fig:attntionmaps} visualizes the attention maps. We also notice the ``end'' attention map $M^{*}_{\text{end}}$ mostly corresponds to noise in other maps, inspiring us to perform a weighted sum of these attention maps for an aggregated attention map at the first stage as:
\begin{equation}\label{eq:aggregated}
M_{1}^{*}=mean(\sum_{i\in token}M^{*}_{i}-\gamma \cdot M_{\text{end}}^{*}),
\end{equation}
where $\gamma$ represents the strengths of the noise-reduced attention maps.
The aggregated cross-attention map $M_{1}^{*}$ highlights the text and surrounding areas. To capture the rough text areas, we perform 3-means clustering on $M_{1}^{*}$, selecting the top two categories with higher pixel brightness as the mask areas. This can be expressed as:
\begin{equation}
\begin{split}
C_{1}^{1}, \;&C_{1}^{2}, \;C_{1}^{3}=3\text{-}means({M}_{1}^{*}),\\
\mathcal{M}_1(i,j)&=
    \begin{cases} 
    1 & \text{if } {M}_{1}^{*}(i,j) \in C_{1}^{1}\cup C_{1}^{2}, \\
    0 & \text{otherwise,} 
    \end{cases}
\end{split}
\end{equation}
where $C_{1}$, $C_{2}$, and $C_{3}$ denote the three clustering results, ordered by pixel values from high to low. $\mathcal{M}_1$ is a rough mask capturing the text regions as illustrated in Fig.~\ref{fig:framework}. If user provides a mask for fine or selective text destruction, it can be directly treated as $\mathcal{M}_1$.


\subsubsection{Continuous Text Adjustment}

After obtaining an introductory text region mask, we use a similar process to continuously adjust the mask. We treat each connected component in the mask $\mathcal{M}_1$ as an isolated text region and crop $n$ corresponding sub-images, $\{\boldsymbol{x}^{1}_{0},\boldsymbol{x}^{2}_{0},\dots,\boldsymbol{x}^{n}_{0}\}$, from the original image $\boldsymbol{x}_{0}$.
Following on, we carry out $n$ inversion processes, each further adjusting the text region mask for every cropped image. For the $k$-th sub-image's inversion process, we compute its aggregated attention map according to Eq.~\eqref{eq:aggregated} and then resize it to the sub-image shape, denoted as ${M}_{2,k}^{*}$. To further exclude non-text regions, we handle ${M}_{2,k}^{*}$ using 2-means:
\begin{equation}
\begin{split}
C_{2,k}^{1}, \;C_{2,k}^{2}&=2\text{-}means({M}_{2,k}^{*}),\\
\mathcal{M}_{2,k}(i,j)=&
    \begin{cases} 
    1 & \text{if } {M}_{2,k}^{*}(i,j) \in C_{2,k}^{1}, \\
    0 & \text{otherwise.} 
    \end{cases}
\end{split}
\end{equation}

After $n$ inversion processes, we obtain a set of adjusted masks $\{\mathcal{M}_{2,1},\mathcal{M}_{2,2},\dots,\mathcal{M}_{2,n}\}$, removing more non-text regions. Taking the union of all mask regions yields a refined mask $\mathcal{M}_{2}$. To ensure all text is within the mask, we perform a dilation operation on $\mathcal{M}_{2}$:
\begin{equation}
\mathcal{M}_2=dilation(\mathcal{M}_2,(k_{1},k_{1})),
\end{equation}
where $(k_{1},k_{1})$ denotes the kernel size.
Fig.~\ref{fig:framework} visualizes $\mathcal{M}_2$ which further filters out most non-text regions of $\mathcal{M}_1$.


\subsubsection{Meticulous Text Delineation}
In our continued efforts to meticulously delineate the details of text edges from the image, we adopt a specific approach. For each sub-image $\boldsymbol{x}_0^k$, we engage in 2-means clustering, allowing us to segregate the image into two distinct classes. It's worth noting that during this process, we do not have prior knowledge of which segment represents text and which constitutes the background:
\begin{equation}
\begin{split}
C_{3,k}^{1}, \; C_{3,k}^{2}&={2\text{-}means}(\boldsymbol{x}^{k}_{0}),\\
\mathcal{M}_{\text{tmp},k}^{c}(i,j)=&
    \begin{cases} 
    1 & \text{if } \boldsymbol{x}^{k}_{0}(i,j) \in C_{3,k}^{c}, \\
    0 & \text{otherwise.} 
    \end{cases}
\end{split}
\end{equation}

Let $\mathcal{R} = \mathcal{M}_1 - \mathcal{M}_2$ define the representation of non-text regions that have been filtered through the second stage. By closely examining the distribution ratios of the two clusters within areas that were previously identified as non-text, we can effectively discern and pinpoint the text regions:
\begin{equation}
\mathcal{M}_{3,k}=
\begin{cases} 
    \mathcal{M}_{\text{tmp},k}^{1} & \text{if } sum({\mathcal{R} \cdot  \mathcal{M}_{\text{tmp},k}^{1}}) \le sum({\mathcal{R} \cdot \mathcal{M}_{\text{tmp},k}^{2}}), \\
    \mathcal{M}_{\text{tmp},k}^{2} & \text{otherwise.} 
\end{cases}
\end{equation}

By intersecting $\mathcal{M}_1$ with the union of all clusters that represent text regions, we derive the final text mask $\mathcal{M}_3$:
\begin{equation}
\mathcal{M}_3=\mathcal{M}_1 \cdot \sum_{k=1}^{n}\mathcal{M}_{3,k}.
\end{equation}

Algorithm~\ref{alg:localization_standard} shows the details of our hierarchical text localization and Algorithm~\ref{alg:localization_mask} specifies the version of available mask for fine or selective text destruction.

\subsection{Text Region Destruction}
Although the latent start code $\boldsymbol{z}_{T}^{*}$ generated by inversion of Eq.~\eqref{eq:inversion} adheres to a Gaussian distribution, it is not entirely stochastic, as deterministic inference processes govern this aspect. The text regions are no exception in this respect. Consequently, we must destroy the original latent code of the text regions to prevent their recovery during the denoising process. 
In order to achieve this, we put forward a strategy that involves filling the latent code within the $\mathcal{M}_{3}$ region with fresh random Gaussian noise.

\begin{algorithm}[!h]
    \caption{Text Destruction and Background Restoration}
    \label{alg:destruction_restoration}
    \begin{algorithmic}[1]
        \REQUIRE latent list $Z$ and accurate text mask $\mathcal{M}_{3}$.
        \ENSURE image with the text regions destroyed $\boldsymbol{x}'_{0}$.
        \STATE Initialize a random Gaussian noise $\boldsymbol{\epsilon}$ of size like $Z[T]$
        \STATE  $\boldsymbol{z}_{T} = Z[T]$
        \STATE  $\boldsymbol{z}'_{T} = Z[T]\cdot(1-\mathcal{M}_{3}) + \boldsymbol{\epsilon}\cdot\mathcal{M}_{3}$
        \FOR{$t = T$ to $1$}
            \IF{$t$ == 2}
                \STATE $\mathcal{M}_{3}'= \text{dilation} (\mathcal{M}_{3},(k_2,k_2))$
                \STATE $\boldsymbol{z}_t'=\boldsymbol{z}_t\cdot(1-\mathcal{M}_{3}') + \boldsymbol{z}_t'\cdot\mathcal{M}_{3}'$
            \ENDIF
            \STATE $\boldsymbol{\epsilon},K,V= \text{denoising}_{\theta} (\boldsymbol{z}_t,t)$
            \IF{t in $[1 \dots 45]$}
                \STATE $\boldsymbol{\epsilon}'= \text{denoising\_KV\_combination}_{\theta} (\boldsymbol{z}_t',t,K,V,\mathcal{M}_{3})$
            \ELSE
                \STATE $\boldsymbol{\epsilon}'= \text{denoising}_{\theta} (\boldsymbol{z}_t',t)$
            \ENDIF
            \STATE $\boldsymbol{z}_{t-1} = 
            Z[t-1]$
            \STATE $\boldsymbol{z}_{t-1}' = 
\sqrt{\frac{\alpha_{t-1}}{\alpha_{t}}} \boldsymbol{z}_{t}' + 
(\sqrt{\frac{1}{\alpha_{t-1}}-1}-\sqrt{\frac{1}{\alpha_{t}}-1})\boldsymbol{\epsilon}'$
        \ENDFOR
        \STATE $\boldsymbol{x}'_{0}=\text{autodecoder}(\boldsymbol{z}'_{0})$
    \end{algorithmic}
\end{algorithm}

The characteristics of this noise, specifically its mean $\mu$ and variance $\sigma$, are not arbitrarily chosen. Instead, they are calculated based on the entire latent. This approach ensures that the noise introduced aligns with the overall distribution of the latent code, maintaining the integrity of the data while still disrupting the original latent code of the text regions. This method is detailed as follows:
\begin{equation}
    \begin{split}
    \mu=\frac{1}{hw}&\sum_{i=1}^{h}\sum_{j=1}^{w}{z}_{T}^{*}(i,j),\;
    \sigma=\frac{1}{hw}\sum_{i=1}^{h}\sum_{j=1}^{w}({z}_{T}^{*}(i,j)-\mu)^2,\\
    z’_{T}=&(1-\mathcal{M}_3) \cdot \*{z}_{T}^{*}+
    \mathcal{M}_3 \cdot \epsilon, \quad \epsilon \sim \mathcal{N}
    (\mu\cdot\boldsymbol{I}, \sigma\cdot\boldsymbol{I}),
    \end{split}
\end{equation}
where the newly $z’_{T}$ denotes the destroyed latent code.

\subsection{Non-Text Region Restoration}

In denoising $z'_T$, we also expect to well restore the non-text region.
Inspired by Cao~\textit{et al.}~\cite{cao2023masactrl}, we recognize the role of the key $K$ and value $V$ in providing robust guidance for the structure and texture of specific objects. We employ a process where we extract $K^*$ and $V^*$ from the self-attention layer of the denoising procedure for the source image $\boldsymbol{x}_0$. 
We then inject these into $K'$ and $V'$ during denoising $z'_T$ in accordance with the mask $\mathcal{M}_3$, ensuring well object restoration:
\begin{equation}\label{eq:kv-combination}
    \begin{split}
    K'=K^* \cdot (1-\mathcal{M}_3) + K' \cdot \mathcal{M}_3,\\
    V'=V^* \cdot (1-\mathcal{M}_3) + V' \cdot \mathcal{M}_3.
    \end{split}
\end{equation}

To avoid reintroducing text into the background reconstruction, we limit this operation to specific denoising steps and self-attention layers. As manifested in Fig.~\ref{fig:framework}, we use the $KV$ combination from $t = 45 \rightarrow 0$ steps at one self-attention layer of the U-Net's front end and two layers at the back end.

To mitigate the errors between the reconstructed image and the original one, we save the latent at each inversion step and replace it during source image reconstruction at each denoising step. We also use latent code replacement for explicit background restoration. At the $t=2$ step, we reintroduce the source latent of non-text regions into the background for the matching with the original image:
\begin{equation}\label{eq:replacement}
    \begin{split}
    \mathcal{M}_3=dilation(\mathcal{M}_3,(k_{2},k_{2})),\\
    z’_{2}=z^*_{2} \cdot (1-\mathcal{M}_3)+z'_{2} \cdot \mathcal{M}_3.
    \end{split}
\end{equation}

A more detailed text destruction and background restoration process are demonstrated in Algorithm\,\ref{alg:destruction_restoration}.

\section{Experimentation}

\subsection{Experimental Setups}
We use the pre-trained stable diffusion 1.5 model~\cite{stable-diffusion-1.5} to validate our TextDestroyer approach.
During both the denoising and inversion processes, we employ the DDIM sampling strategy~\cite{song2020denoising}, encompassing 50 steps.
In the attention aggregation phase, we set $\gamma$ to $1.5$.
For mask dilation, we designate the kernel sizes as $k_1  = 5$ and $k_2 = 9$.
When a user-provided mask is available, we not only replace it with $\mathcal{M}_1$, but also forgo the use of dilated $\mathcal{M}_3$, opting instead to employ it in identifying the areas that require restoration during the latent code replacement process.

We compare our method with EraseNet~\cite{liu2020erasenet}, MTRNet~\cite{tursun2019mtrnet}, GaRNet~\cite{lee2022surprisingly}, STRDD~\cite{yang2022strdd}, DeepEraser~\cite{feng2024deeperaser}, and CTRNet~\cite{liu2022don}. Experiments are conducted on SCUT-Enstext~\cite{zhang2019ensnet}, as well as through a qualitative analysis on generated images.

\subsection{Evaluation Metrics}
As per convention, our method is assessed by: similarity evaluations and detection evaluations.
Similarity evaluations aim to gauge the resemblance between the output image and the ground-truth image, thereby quantifying the effectiveness of background restoration.
We use the PSNR and the MSSIM for assessment.
Detection evaluations focus solely on measuring the extent of text erasure, without considering the quality of background restoration.
Consistent with prior research~\cite{liu2020erasenet,liu2022don,lee2022surprisingly,yang2022strdd,feng2024deeperaser}, we employ CRAFT~\cite{baek2019character} as the text detector to compute recall (R), precision (P), and F-score (F).

\subsection{Performance Analysis}

\begin{figure}[!t]
    \centering
    \includegraphics[width=0.45\textwidth]{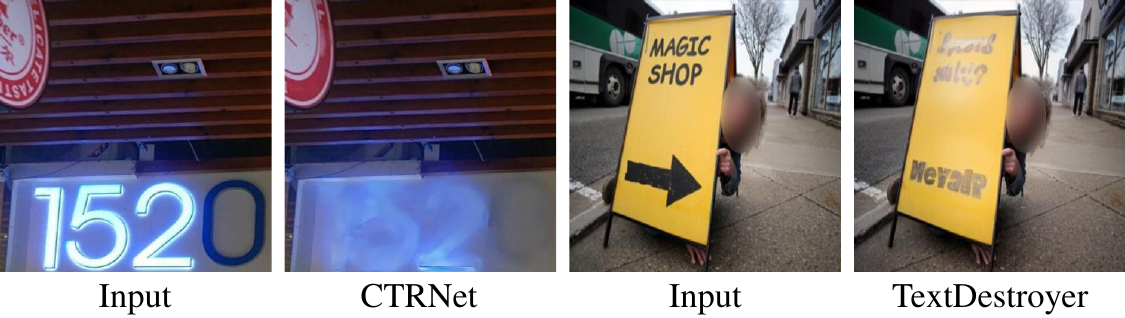}
    \caption{Visual comparison of typical failures between scene text removal method CTRNet and TextDestroyer.}
    \label{fig:fails}
\end{figure}

\subsubsection{Quantitative Comparison}
\begin{table}[!h]
\centering
\caption{Quantitative comparison on SCUT-EnsText. The $\dagger$ denotes text area masks required, and $^*$ indicates re-implemented models due to unavailability of the source.}
\label{tab:quantitative}
\begin{tabular}{cccccc}
\toprule
Methods& PSNR$\uparrow$& MSSIM$\uparrow$& P$\downarrow$& R$\downarrow$& F$\downarrow$\\
\midrule
EraseNet& 35.87& 98.65& 75.5& 99.0& 85.7\\
MTRNet$\dagger$& 25.72& 95.2& 73.8& 98.9& 73.8\\
GaRNet$\dagger$& 36.79& 99.31& 26.4& 65.0& 37.6\\
STRDD${\dagger }{}^*$& 34.84& 94.75& -& 4.6& -\\
Deeperaser$\dagger$& 36.67& 99.33& 4.2&  36.3&  7.5\\
CTRNet$\dagger$& 36.82& 99.24& 1.4& 0.0& 0\\
TextDestroyer (mask)$\dagger$& 30.06& 96.22& 41.5& 89.9& 56.9\\
TextDestroyer& 17.11& 77.12& 46.4& 94.8& 62.36\\
\bottomrule
\end{tabular}
\end{table}

Table~\ref{tab:quantitative} shows quantitative results on SCUT-Enstext~\cite{zhang2019ensnet}.
Despite relying solely on pre-trained models for tasks like text localization and removal, our performance still exhibits a certain gap compared to state-of-the-art (SOTA) models~\cite{liu2020erasenet, tursun2019mtrnet, lee2022surprisingly, yang2022strdd, feng2024deeperaser, liu2022don, zhang2019ensnet, zhang2019ensnet}.
For a specific analysis, detection evaluations merely confirm the presence of a distribution resembling text in the image, regardless of whether recognizable traces exist.
Off-the-shelf text erasure methods often erase text to a sufficient depth to deceive character detection models.
In contrast, our TextDestroyer completely destroys the text.
In our failure cases of Fig.~\ref{fig:fails}, the residual distribution of shapes resembling text mostly contains unrecognizable information.
Moreover, the comparison models are all trained with annotations and tested on the same dataset, achieving high scores without requiring strong generalization capabilities.
In summary, our method diverges from scene text erasure methods in terms of motivation, technical approach, and evaluation criteria.
\textit{Evaluation on traditional scene text removal datasets fails to accurately gauge the effectiveness of our method}.
Thus, quantitative experiments serve as a reference only, while qualitative experiments will further illustrate our superiority.


\subsubsection{Qualitative Comparisons}

\begin{figure*}[!t]
    \centering
    \includegraphics[width=.92\textwidth]{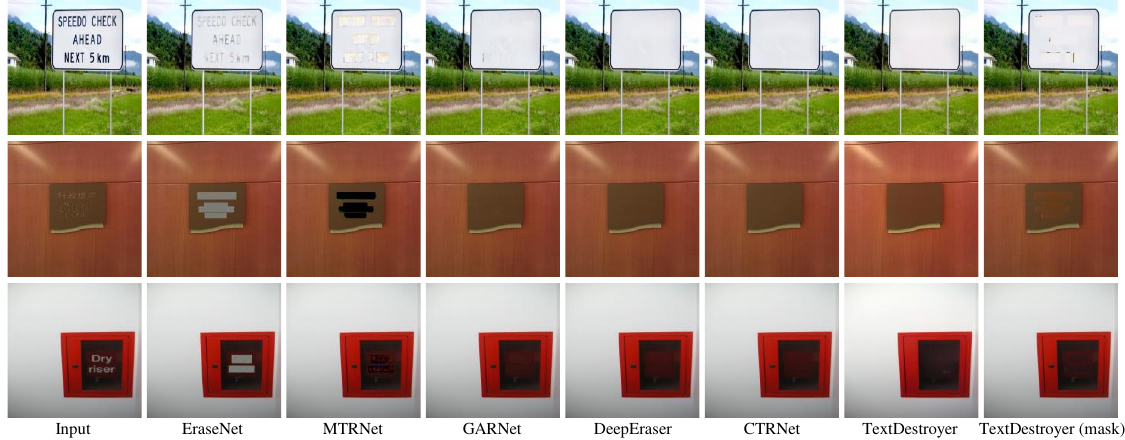}
    \caption{Visual comparison of TextDestroyer and other scene text removal methods on the SCUT-Enstext dataset.}
    \label{fig:enstext_images}
\end{figure*}

\begin{figure*}[!t]
    \centering
    \includegraphics[width=.92\textwidth]{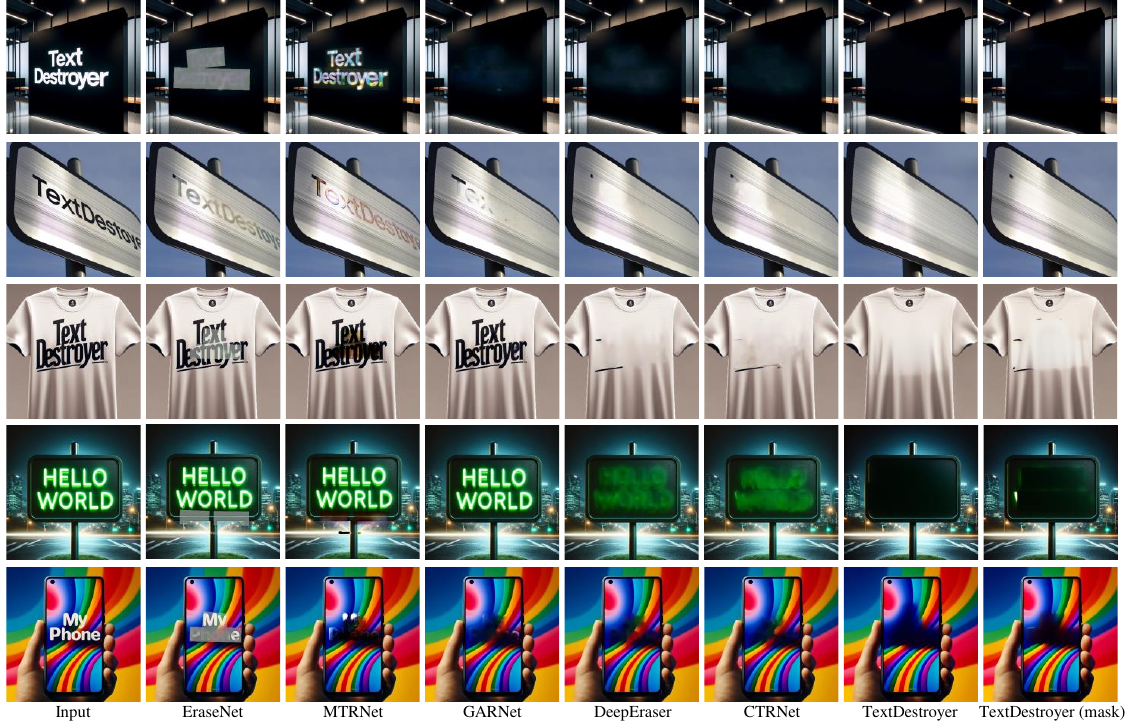}
    \caption{Visual comparison of TextDestroyer and other scene text removal methods on generated images.}
    \label{fig:generated_images}
\end{figure*}

In Fig.~\ref{fig:enstext_images}, we commence by evaluating the performance of TextDestroyer against various alternative techniques on the SCUT-Enstext dataset.
As elucidated in the preceding section, the qualitative outcomes presented herein serve solely as a point of reference.

\begin{figure*}[!t]
    \centering
    \includegraphics[width=.92\textwidth]{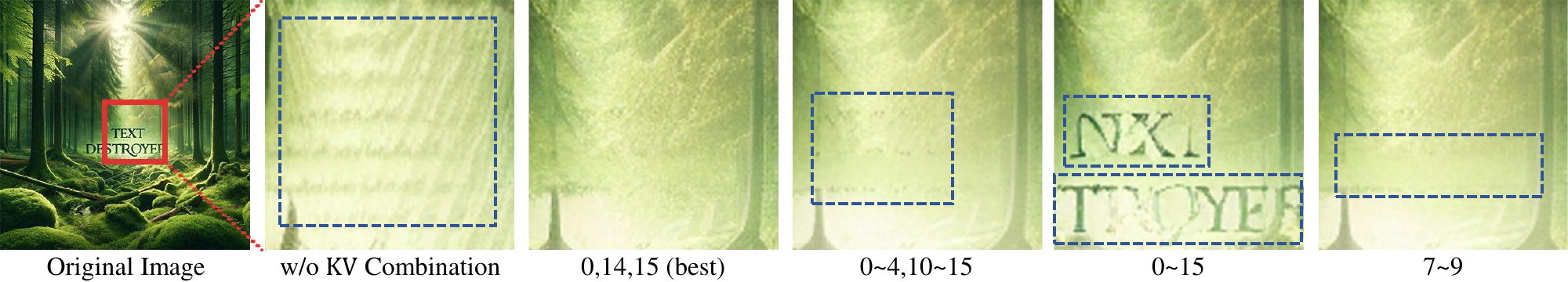}
    \caption{Visual results of different $KV$ combination layers. The numbers below the image indicate the identifiers of self-attention layers where $KV$ combination occurs. Defective areas are marked with blue dashed lines. Best view with zooming in.}
    \label{fig:ablation_kv_layers}
\end{figure*}
\begin{figure*}[!t]
    \centering
    \includegraphics[width=.92\textwidth]{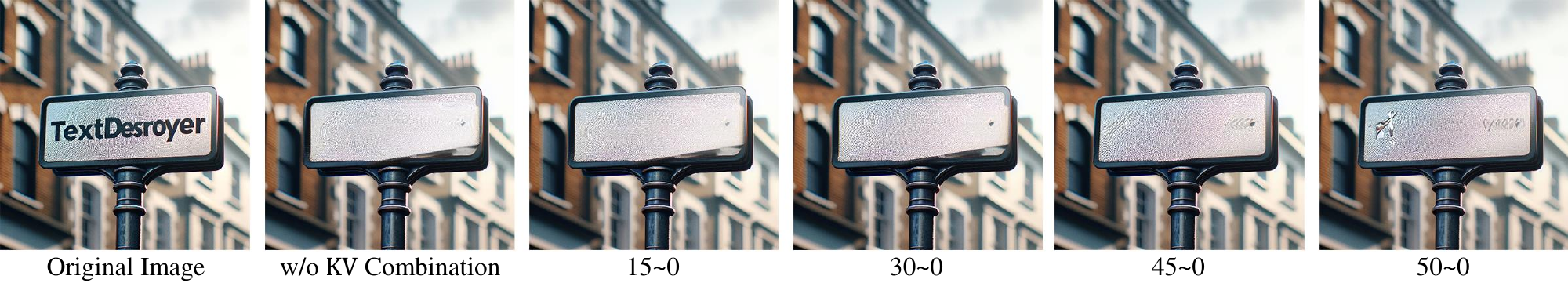}
    \caption{Visual results of $KV$ combination steps. The numbers below image indicate the denoising steps with $KV$ combination. Best view with zooming in.}
    \label{fig:ablation_kv_steps}
\end{figure*}

Fig.~\ref{fig:generated_images} juxtaposes the efficacy of our approach on synthetic images generated by ``DALL-E 3''~\cite{betker2023improving} with that of other methodologies.
The assortment of generated images exhibits a greater diversity in comparison to authentic photographs.
Throughout the creation process, users possess the autonomy to tailor the imagery to their preferences by employing descriptive prompts, encompassing real-world vistas, animated styles, and abstract art forms, among others.
It becomes evident that managing synthetic images necessitates a robust capacity for generalization.
Conventional endeavors in erasing text from images have predominantly relied on the incremental fading of textual elements within networks, a technique that may leave perceptible remnants in the resultant images.
While some of these vestiges are discernible upon cursory inspection, others, though less conspicuous, can be identified upon magnification and meticulous scrutiny, thereby disclosing the original text.
In contrast to our innovative TextDestroyer, competing text removal solutions face challenges in accurately replicating intricate textural details.
These methods often prioritize the preservation of low-frequency background information at the expense of dedicated high-frequency textures, leading to a more natural visual outcome, albeit with a potential trade-off in terms of similarity metrics.
Moreover, as the transition is made from specific datasets to generative imagery, traditional text removal strategies, even when applied to images closely resembling real-world scenes, are prone to a heightened likelihood of error, indicative of significant overfitting.

\subsection{Ablation Studies}
In this section, we conduct extensive ablation experiments to examine the characteristics of our model components, including hierarchical text localization in Sec.~\ref{sec:localization}, $KV$ combination in Eq.~\eqref{eq:kv-combination} and latent code replacement in Eq.~\eqref{eq:replacement}.

\begin{table}[!h]
\centering
\caption{Quantitative ablation study.}
\label{tab:ablation}
\begin{tabular}{cccccc}
\toprule
Methods& PSNR$\uparrow$& MSSIM$\uparrow$& P$\downarrow$& R$\downarrow$& F$\downarrow$\\
\midrule

TextDestroyer& \textbf{17.11}& \textbf{77.12}& 46.4& 94.8& 62.4\\
\cline{1-6}
w/o KV Combination& 12.6& 42.1& \textbf{11.8}& \textbf{65.0}& \textbf{20.0}\\
KV layer 0-4,10-15& 13.5& 64.3& 62.2& 97.4& 75.9\\
KV layer 0-15& 13.8& 66.3& 62.6& 97.8& 76.3\\
KV layer 7-9& 13.1& 60.9& 46.0& 93.8& 61.7\\
KV step 50-0& 12.4& 51.2& 34.3& 90.1& 49.7\\
KV step 15-0& 12.3& 43.8& 17.8& 78.8& 29.0\\
\cline{1-6}
w/o latent replace& 12.3& 50.7& 35.5& 90.4& 60.0\\
latent replace at t=30& 12.8& 57.8& 42.1& 93.2& 58.0\\
replace after decoder& 14.3& 70.4& 50.7& 95.6& 66.3\\
\bottomrule
\end{tabular}
\end{table}

\subsubsection{Steps and Layers for $KV$ Combination}

Subsequently, we conduct ablation studies on the layers where the $KV$ combination is employed.
To achieve a more pronounced effect, during the visual ablation process, we do not replace the latent code for the diffusion providing $KV$, but in the quantitative ablation, we maintain the standard settings.
As depicted in Fig.~\ref{fig:ablation_kv_layers}, an excessive number of $KV$ combination layers leads to the re-emergence of textual artifacts within the generated images.
Insufficient $KV$ comparison layers result in the retention of areas disrupted by random noise, thereby compromising the visual fidelity.
When the $KV$ combination layers are fixed and positioned closer to the middle, owing to limitations in resolution, their capacity to restore background textures diminishes, consequently favoring the generation of smoother, blurred textures.
We have also implemented ablation on the step where the $KV$ combination is employed.
Similar to the ablation on $KV$ combination self-attention layers, an excessive number of $KV$ combination steps leads to text reappearance, while too few result in background disorder.
In Fig.~\ref{fig:ablation_kv_steps}, during this process of change, there might not exist a perfect point where the background is completely restored while also not introducing information from the text area.

As shown in the second data block of Table~\ref{tab:ablation}, the phenomena mentioned above are also validated in quantitative results: a lack of KV combination may lead to a chaotic background, such that although text-like distributions cannot be detected by detection metrics, PSNR and MSSIM are reduced due to the unreliable background; excessive or inappropriate KV combination reintroduces text into the background, resulting in low PSNR and MSSIM and high detection metrics. Notably, even in cases where the text distribution is recognized by the detection model, as shown in Fig.~\ref{fig:fails}, the residual traces remain unreadable. This means we can ensure that the text is thoroughly obliterated with a degree of redundancy.

After numerous experiments, we have determined that optimal outcomes are attained when the $KV$ combination is executed within $t=45\rightarrow0$ denoising steps, specifically within the first and the last two self-attention layers of the U-Net architecture.

\subsubsection{Stages of Hierarchical Text Localization}
\begin{figure}[!t]
    \centering
    \includegraphics[width=0.45\textwidth]{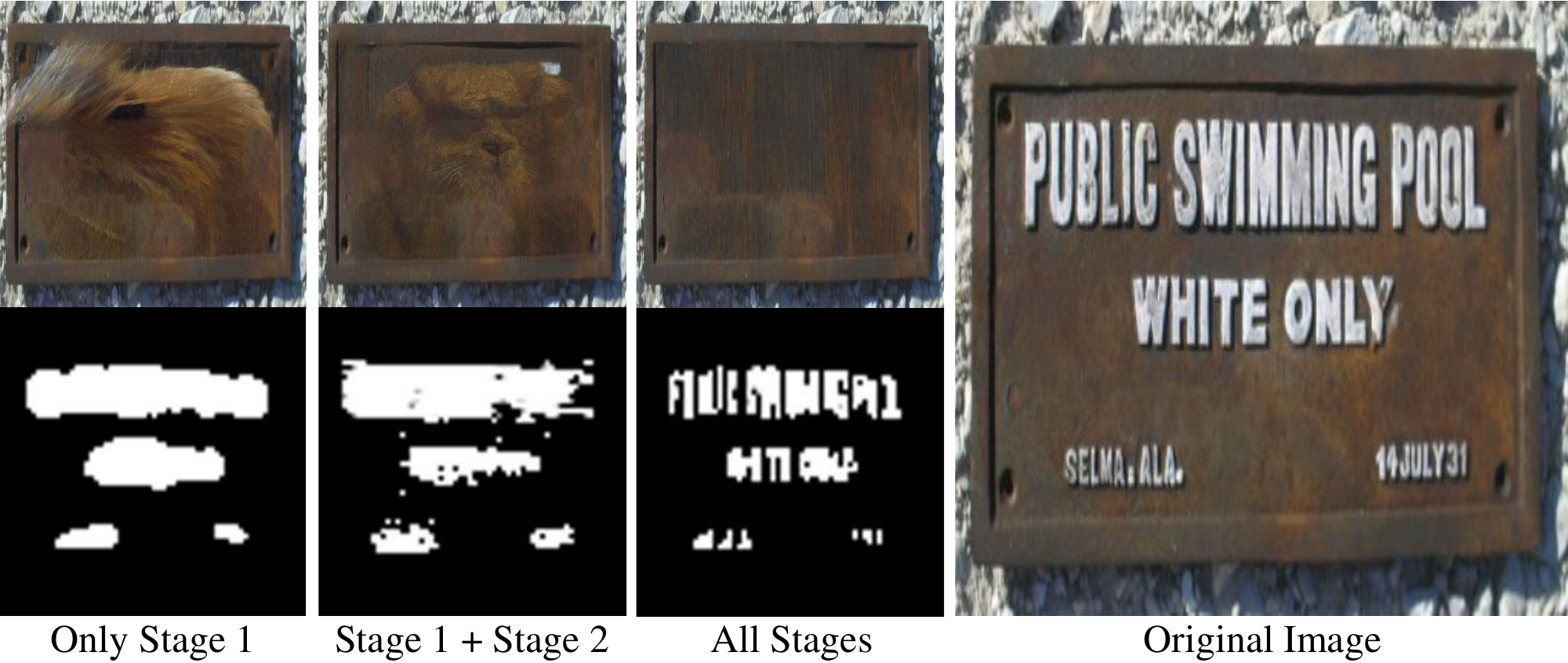}
    \caption{Visual results of hierarchical text localization stages. Best view with zooming in.}
    \label{fig:ablation_local_stages}
\end{figure}

To optimize the precision of token localization within the diffusion process, we have meticulously crafted a hierarchical text localization strategy that unfolds in as three-tiered sequence.
Since masks from three stages have different optimal parameter choices when applied individually in subsequent steps, especially the size of dilation kernels, isolating a mask from any single stage results in biased metrics. Therefore, we only provide an intuitive display on the visual effects.
As delineated in Fig.~\ref{fig:ablation_local_stages}, there is a progressive refinement in the accuracy of the mask as the process nears completion.
Concurrently, the image that emerges above exhibits enhanced visual fidelity and plausibility.
The mask encapsulating the textual region is progressively refined, ensuring that the background is effectively excluded while meticulously preserving the integrity of the text area, ultimately converging to the text's edge.
Employing the comprehensive three-stage text localization technique yields images that are visually superior and more convincingly devoid of textual artifacts.

\subsubsection{Step for Latent Code Replacement}

\begin{figure}[!t]
    \centering
    \includegraphics[width=0.45\textwidth]{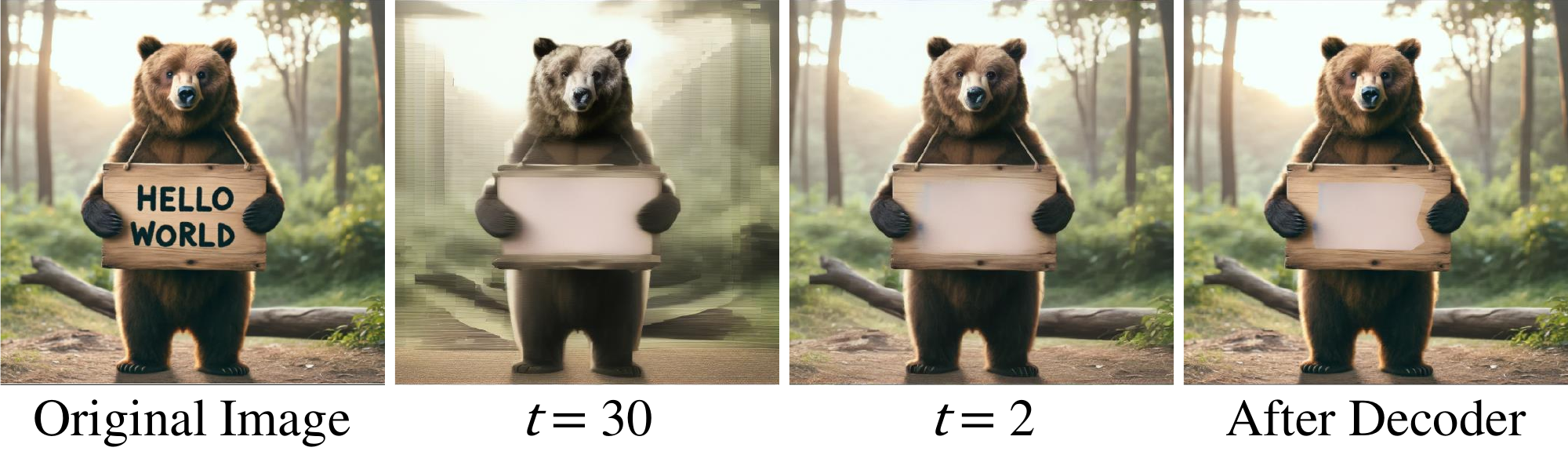}
    \caption{Visual results of latent code replacement steps. Best view with zooming in.}
    \label{fig:ablation_latent}
\end{figure}

The objective of latent code replacement is to rectify the damage incurred by the background during the process of approximate inversion and the ensuing disrupted reconstruction.
As depicted in Fig.~\ref{fig:ablation_latent} and the third block of Table~\ref{tab:ablation}, the premature introduction of the source latent code into the denoising reconstruction process leads to a compromised restoration efficacy.
On the other hand, the delayed introduction of the latent code may give rise to a perceptible discontinuity at the boundary between the eradicated text area and the surrounding background.
In light of the delicate balance that must be struck between the thorough restoration of the background and the seamless integration of the latent code, we have strategically selected to execute the latent code replacement at the time step denoted as $t=2$.
This decision is informed by a nuanced consideration of both the quality of background restoration and the continuity of the image's composition.

\subsection{Limitations and Discussions}

Our TextDestroyer faces certain challenges: (1) the refinement in restoring background regions is lacking, occasionally leading to inaccuracies in color, texture, and structure; (2) it struggles with curved and small text due to limitations of the pre-trained model; (3) the inference process is time-consuming, taking approximately 25 to 60 seconds per image on a single 3090 GPU. These limitations imply that our future research could focus on using pre-trained models with enhanced text detection capabilities to streamline text localization and thus expedite inference. Moreover, improving the robustness of text localization and the quality of background restoration are areas that merit further exploration.

\section{Conclusion}

In this study, we have presented TextDestroyer, a novel model for destroying scene text without training or annotation. Using a pre-trained diffusion model, TextDestroyer achieves hierarchical text localization and introduces random noise to disrupt text distribution. The denoising phase employs $KV$ combination and latent code replacement for background restoration. Our approach differs from others by fully destroying text distribution before reconstructing the background, avoiding residual traces. Experiments show our proposed TextDestroyer excels at complete text removal and provides enhanced generalization for varied inputs, creating realistic backgrounds rather than just smoothing.

\section{Acknowledgments}
\noindent This work was supported by National Science and Technology Major Project (No. 2022ZD0118202), the National Science Fund for Distinguished Young Scholars (No.62025603), the National Natural Science Foundation of China (No. U21B2037, No. U22B2051, No. U23A20383, No. 62176222, No. 62176223, No. 62176226, No. 62072386, No. 62072387, No. 62072389, No. 62002305 and No. 62272401), and the Natural Science Foundation of Fujian Province of China (No.2022J06001).

\bibliographystyle{unsrt}  
\bibliography{templateArxiv}

\end{document}